\documentclass[runningheads]{llncs}
\usepackage[T1]{fontenc}
\usepackage{amsmath}
\usepackage{amssymb}
\usepackage{graphicx}
\usepackage{booktabs}
\usepackage{multirow}

\begin{document}
\title{Conv-like Scale-Fusion Time Series Transformer: A Multi-Scale Representation for Variable-Length Long Time Series}
\titlerunning{Conv-like ScaleFusion Time Series Transformer}
%
\author{Kai Zhang\inst{1} \and Siming Sun\inst{1} \and Zhengyu Fan\inst{1} \and Qinmin Yang\inst{1} \and Xuejun Jiang\inst{1}}
\authorrunning{Kai Zhang et al.}
%
%
\institute{Zhejiang University, Hangzhou 310027, China}
\maketitle              
\begin{abstract}
    Time series analysis faces significant challenges in handling variable-length data and achieving robust generalization. While Transformer-based models have advanced time series tasks, they often struggle with feature redundancy and limited generalization capabilities. Drawing inspiration from classical CNN architectures' pyramidal structure, we propose a Multi-Scale Representation Learning Framework based on a Conv-like ScaleFusion Transformer. Our approach introduces a temporal convolution-like structure that combines patching operations with multi-head attention, enabling progressive temporal dimension compression and feature channel expansion. We further develop a novel cross-scale attention mechanism for effective feature fusion across different temporal scales, along with a log-space normalization method for variable-length sequences. Extensive experiments demonstrate that our framework achieves superior feature independence, reduced redundancy, and better performance in forecasting and classification tasks compared to state-of-the-art methods.


\end{abstract}
\section{Introduction}
Time series (TS) data are foundational across numerous domains, such as acoustics, healthcare, and manufacturing. While deep learning models have achieved remarkable progress in time series forecasting \cite{liu2022non,wu2021autoformer,wu2022timesnet,zhou2022fedformer,nie2022time,wang2024timemixer}, classification, and anomaly detection, their generalization capabilities remain suboptimal \cite{zerveas2021transformer,lim2021temporal}. This limitation is primarily due to the dominance of Transformer-based architectures, which typically extract features at the granularity of time entries—where a time entry may correspond to a single time point, a time window, or a patch. As a result, the extracted features are often high-dimensional. Subsequently, a task-specific head is employed to flatten these temporally structured features for downstream tasks such as prediction or classification. However, in practice, the encoded features of time series data frequently exhibit significant redundancy, particularly when processing long sequences. Directly leveraging these features for downstream tasks can cause the model to overemphasize redundant information, thereby compromising its generalization performance. Furthermore, time series data collected in real-world scenarios often have variable lengths. This also leads to inconsistent feature dimensions after feature extraction by traditional Transformer-based methods, making unified task learning for variable-length time series data challenging. Therefore, efficiently representing time series data while accommodating variations in input length remains a critical challenge in time series analysis.

While Transformer-based time series models utilize the attention mechanism to reduce the effective distance between features at each time entry to one—thereby enabling rapid global feature aggregation—this strength has inadvertently led researchers to overlook a crucial strategy from the era of convolutional neural networks (CNNs). Classical models such as VGG\cite{simonyan_very_2015}, ResNet backbones\cite{he_deep_2016}, and architectures like FPN\cite{lin_feature_2017} or U-Net\cite{ronneberger_u-net_2015} adopt a characteristic pyramidal structure, in which each downsampling operation doubles the number of channels. By only moderately decreasing the spatial resolution of feature maps and distributing independent information across an increased number of feature channels, these models facilitate efficient feature fusion from local to global scales, ultimately resulting in highly effective data representations. Furthermore, we observe that in such feature extraction processes, each convolutional layer operates on features at different physically meaningful scales. This insight provides a promising direction for addressing the challenge of representing variable-length time series data.

Therefore, we propose a Multi-Scale Representation Learning Framework based on a Conv-like ScaleFusion Transformer to address variable-length long time series. First, we redefine a temporal convolutional structure by combining patching operations with multi-head attention, enabling progressive compression of the temporal dimension and expansion of feature channels through a novel sequence of temporal convolutions. This design facilitates the extraction of features at multiple temporal scales. Second, we introduce a novel cross-scale attention mechanism to replace the traditional residual structure, enabling effective cross-layer fusion of features from different temporal scales. Finally, we propose a log-space normalization method for variable-length time series features, ensuring that sequences of different lengths can ultimately yield multi-scale feature representations. Experimental results demonstrate that our framework achieves stronger feature independence and reduced redundancy in the representation of long, variable-length time series across multiple datasets, and delivers superior performance on forecasting and classification tasks compared to state-of-the-art counterparts.

Our main contribution can be summarized as follows:
\begin{itemize}
    \item We propose a novel Conv-like ScaleFusion Transformer framework for variable-length long time series, which introduces a temporal convolutional structure that combines patching operations with multi-head attention, enabling progressive compression of the temporal dimension and expansion of feature channels through a novel sequence of temporal convolutions.
    \item We introduce a novel cross-scale attention mechanism to replace the traditional residual structure, enabling effective cross-layer fusion of features from different temporal scales.
    \item We propose a log-space normalization method for variable-length time series features, ensuring that sequences of different lengths can ultimately yield multi-scale feature representations.
\end{itemize}

\section{Related Work}

\subsection{Time Series Representation Learning}

Early neural approaches relied on recurrent (RNN/LSTM) or convolutional (TCN) structures, but the past three years have been dominated by Transformer-style architectures that explicitly target long-range temporal dependencies and cross-variable interactions. Zerveas \emph{et al.}~\cite{zerveas2021transformer} first demonstrated that a vanilla Transformer (TST) trained with masked-reconstruction objectives could serve as a universal encoder for multivariate series. Subsequent work focused on scaling efficiency and inductive biases. Informer~\cite{zhou2021informer} introduced probabilistic sparse self-attention, while Autoformer~\cite{wu2021autoformer} and FEDformer~\cite{zhou2022fedformer} decomposed series into trend/frequency components to improve long-horizon stability. TimesNet~\cite{wu2022timesnet} recast temporal dynamics as 2-D variation patterns, and PatchTST~\cite{nie2022time} showed that non-overlapping patches with local attention outperform point-wise tokens. More recently, iTransformer~\cite{liu2023itransformer} and the Non-Stationary Transformer~\cite{liu2022non} proposed invertible and shift-aware mechanisms to model distributional drift, whereas TimeMixer~\cite{wang2024timemixer} and TimeGPT~\cite{signorini2023timegpt} attempt to build foundation-level models that generalise across domains via large-scale pre-training. Parallel to supervised forecasting, self-supervised objectives such as contrastive TS2Vec~\cite{yue2022ts2vec} and temporal neighbourhood coding (TNC)~\cite{franceschi2019tnc} have proved effective for downstream classification and anomaly detection, underscoring the importance of task-agnostic representation learning. Collectively, these advances highlight a clear trend toward scalable, domain-general encoders that combine efficient attention, frequency reasoning, and self-supervision to capture rich temporal semantics.

\subsection{Transformer-based Time Series Models}

The Transformer architecture initially achieved significant breakthroughs in the field of natural language processing and was rapidly adopted for time series modeling tasks. Compared to traditional recurrent neural networks (RNNs), Transformers are capable of capturing long-range dependencies via the self-attention mechanism, thereby substantially enhancing the model's ability to represent global information. Zerveas et al.~\cite{zerveas2021transformer} were the first to apply the Transformer to general time series representation learning, achieving outstanding performance in both classification and regression tasks. Following this, a series of variants—including Autoformer~\cite{wu2021autoformer}, Informer~\cite{zhou2021informer}, FEDformer~\cite{zhou2022fedformer}, and TimesNet~\cite{wu2022timesnet}—have introduced structural innovations for time series forecasting and anomaly detection, such as adaptive decomposition, sparse attention, and frequency-domain modeling. Despite the strong modeling capabilities of Transformers for time series data, existing approaches primarily focus on feature extraction at the time point or patch level, which limits their ability to fully exploit the multi-scale structural information inherent in time series. Moreover, when dealing with extremely long or variable-length sequences, Transformer models often encounter challenges such as feature redundancy and limited generalization. Therefore, integrating the multi-scale feature extraction strengths of convolutional networks with the global modeling capabilities of Transformers has become a critical research direction in the field of time series modeling.

\section{Proposed Method}

\begin{figure}
  \centering
  \includegraphics[width=0.8\textwidth]{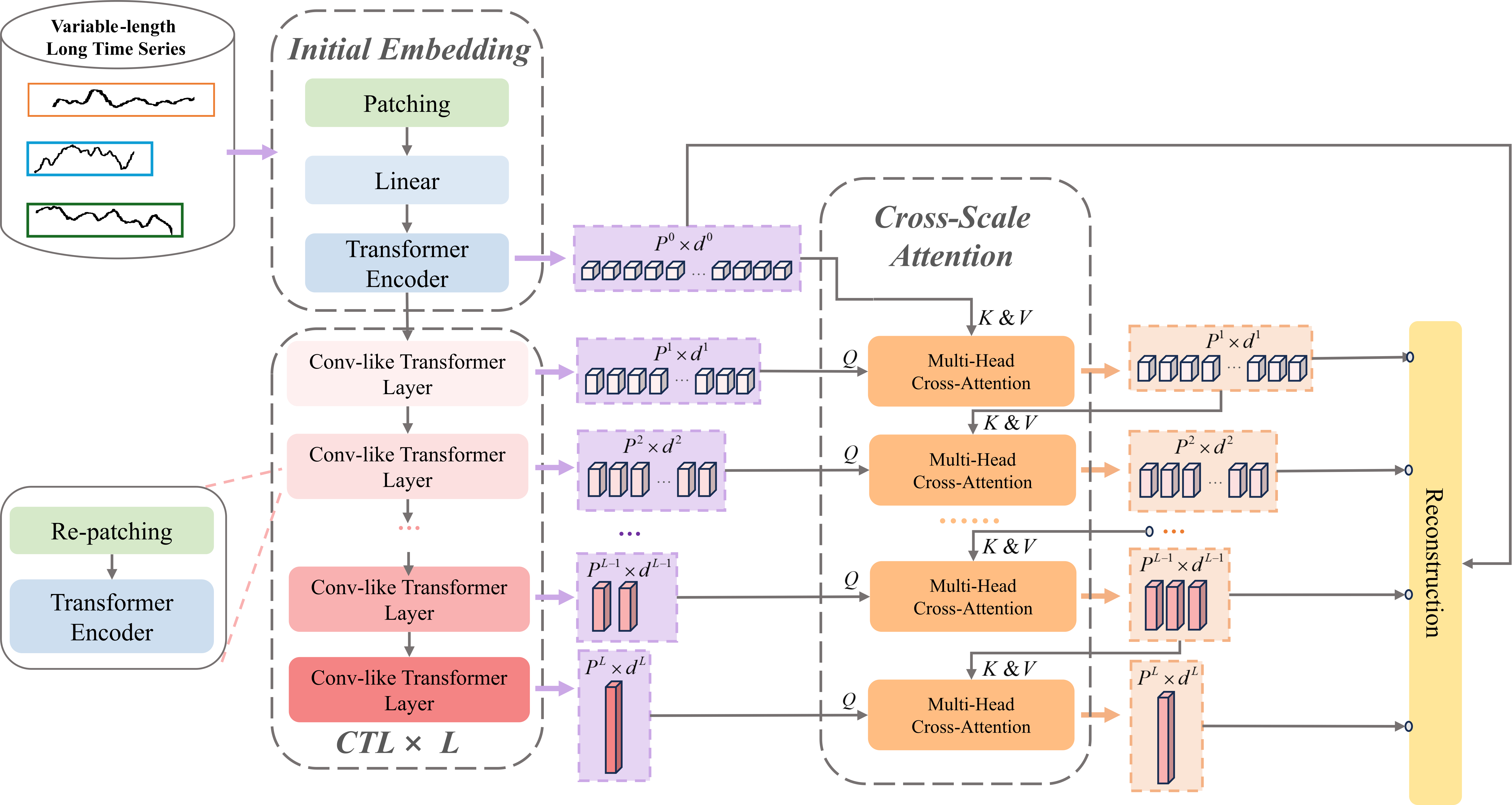}
  \caption{The proposed Conv-like ScaleFusion Time Series Transformer framework.}
  \label{fig:framework}
\end{figure}

The proposed Conv-like ScaleFusion Time Series Transformer framework follows a hierarchical feature extraction process: First, it performs initial embedding by patching the input time series and applying a transformer encoder layer. Then, through a series of Conv-like Transformer Layers (CTLs), each layer progressively compresses the temporal dimension while expanding feature channels, similar to CNN's pyramidal structure. A novel cross-scale attention mechanism is introduced to facilitate feature fusion across different temporal scales and mitigate vanishing gradients. Finally, a log-space normalization strategy is employed to handle variable-length sequences, ensuring consistent feature representation across different input lengths. The framework is trained using a combined loss function that includes both reconstruction loss and feature independence loss. Figure~\ref{fig:framework} shows the entire framework of our proposed method.

\subsection{Initial Embedding}

Given a time series $\mathbf{X} = [x_1, x_2, \cdots, x_T]$, we first perform a patching operation with patch length $l_p$ and stride $s_p$, resulting in the original patch sequence $\mathbf{X}_{\text{patch}} = [\mathbf{x}_{\text{patch}}^p]_{p=1}^{P^0}$, where $\mathbf{x}_{\text{patch}}^p = [x_{p\times s_p : p\times s_p + l_p}]$ is the $p$-th patch and $P^0 = \lfloor \frac{T-l_p}{s_p} \rfloor + 1$ denotes the number of patches. Subsequently, an embedding layer is applied to each patch to project the patch of length $l_p$ into a model dimension of $d^0$. The embedding layer is a simple linear transformation. Then, a transformer encoder layer is applied upon the embedded patches to get the initial temporal features $\mathbf{H}_0 \in \mathbb{R}^{d^0 \times P^0}$. This iniitial embedding process can be expressed as follows:

\begin{align}
    \mathbf{H}_0 &= \text{TransformerEncoders}(\text{LayerNorm}(\mathbf{W}_{\text{emb}} \mathbf{X}_{\text{patch}}))
\end{align}
where $\mathbf{W}_{\text{emb}} \in \mathbb{R}^{d^0 \times l_p}$ is the embedding matrix.

\subsection{Conv-like Transformer}

Each Conv-like Transformer Layer (CTL) consists of a Transformer Encoder Layer followed by a Re-patching Layer. The Re-patching Layer first takes the temporal features produced by the previous layer and performs a new non-overlapping patching operation along the current patch dimension, using a fixed re-patching length of $l_{\text{rp}}$. As a result, the model dimension increases from $d^l$ to $d^{l+1}$, while the temporal length of the features is further reduced to $P^{l+1} = \lceil \frac{P^l}{l_{\text{rp}}} \rceil$. This operation is analogous to the downsampling process in conventional CNNs. The re-patched features are then passed into the Transformer Encoder Layer, where the multi-head attention (MHA) mechanism preserves the model dimension $d^l$. This process is similar to feature extraction in CNNs; however, unlike CNNs, which have a limited receptive field, the MHA mechanism provides a receptive field that covers the entire sequence. By integrating these two components, each CTL effectively compresses the temporal dimension while expanding the feature dimension of the time series features. Importantly, the temporal length and feature dimension of the resulting features are decoupled, allowing for a more flexible and adaptive feature extraction process for time series data.

In this work, we consistently set $d^{l+1} = l_{\text{rp}} \times d^l$ for all $l$ to maintain the identical feature volume within each layer. The maximum input time series length accepted by the model is set to $T_{\text{max}}$. Consequently, after $L_{\text{max}} = \left\lceil \log_{l_{\text{rp}}}\left( \lfloor \frac{T_{\text{max}} - l_p}{s_p} \rfloor + 1 \right) \right\rceil$ CTLs, the number of patches is reduced to 1, and the model dimension becomes $d^L = (l_{\text{rp}})^L \times d^0$.

It is worth noting that the number of activated CTLs varies according to the input sequence length $T$, with only up to $L = \left\lceil \log_{l_{\text{rp}}}\left( \lfloor \frac{T - l_p}{s_p} \rfloor + 1 \right) \right\rceil$ layers being utilized. This design endows the model with strong capability to handle variable-length inputs. Additionally, during the Re-Patching operation at each layer, if the current number of patches is not divisible by $l_{\text{rp}}$, extra patches are padded at the end of the patch sequence to ensure that no information is lost during the patching process. For the $l$-th CTL, the input is the temporal feature $\mathbf{H}_{l-1} \in \mathbb{R}^{ d^{l-1} \times P^{l-1}}$ from the previous layer, and the output $\mathbf{H}_{l}^* \in \mathbb{R}^{d^l \times P^l}$ can be formulated as follows:

\begin{align}
    \mathbf{H}_{l}^{\text{rp}} &= \begin{cases}
        \text{Re-patching}(\mathbf{H}_{l-1}) & \text{if } P^{l-1} \mid l_{\text{rp}} \\
        \text{Re-patching}(padding(\mathbf{H}_{l-1}, \text{len} = l_{\text{rp}})) & \text{if } P^{l-1} \nmid l_{\text{rp}}
    \end{cases}\\
    \mathbf{H}_{l}' &= \text{InstanceNorm}(\mathbf{W}_{l}^{\text{rp}} \mathbf{H}_{l}^{\text{rp}})\\
    {\mathbf{H}_{l}}^* &= \text{LayerNorm}(\text{TransformerEncoders}(\mathbf{H}_{l}')) 
\end{align}
where $\mathbf{W}_{l}^{\text{rp}} \in \mathbb{R}^{d^{l} \times d^{l-1}}$ is the weight matrix for the re-patching operation, and $\mathbf{H}_{l}' \in \mathbb{R}^{d^{l} \times P^l}$ is the re-patched features.

\subsection{Cross-Scale Attention Mechanism}

Each CTL extracts features at a distinct temporal scale. As the network depth increases, the problem of vanishing gradients becomes more pronounced. To address this, inspired by the residual connections in CNN models, we introduce a novel cross-scale attention mechanism that facilitates cross-layer feature fusion and mitigates the vanishing gradient issue. Specifically, for the temporal features ${\mathbf{H}_{l}}^* \in \mathbb{R}^{d^l \times P^l}$ output by the $l$-th CTL, we employ a Multi-head Cross-Attention mechanism to fuse them with the temporal features $\mathbf{H}_{l-1} \in \mathbb{R}^{ d^{l-1} \times P^{l-1}}$ from the previous layer as follows:
\begin{align}
    \mathbf{Q}_{l}^{h} &= \mathbf{W}_{l}^{Q,h} {\mathbf{H}_{l}}^*, \mathbf{K}_{l}^{h} = \mathbf{W}_{l}^{K,h} \mathbf{H}_{l-1}, \mathbf{V}_{l}^{h} = \mathbf{W}_{l}^{V,h} \mathbf{H}_{l-1} \\
    \mathbf{O}_{l}^{h} &= \mathbf{V}_{l}^{h} \text{Softmax}(\frac{{\mathbf{K}_{l}^{h}}^\top {\mathbf{Q}_{l}^{h}}}{\sqrt{d_{\text{csa}}}}) \\
    \mathbf{O}_{l} &= \mathbf{W}_{l}^{O} \text{concat}(\mathbf{O}_{l}^{1}, \mathbf{O}_{l}^{2}, \cdots, \mathbf{O}_{l}^{H}) \\
    \mathbf{H}_{l} &= \text{LayerNorm}({\mathbf{H}_{l}}^* + \mathbf{O}_{l})
\end{align}
where $\mathbf{W}_{l}^{Q,h} \in \mathbb{R}^{d^{l}_{\text{csa}} \times d^l}, \mathbf{W}_{l}^{K,h} \in \mathbb{R}^{d^{l}_{\text{csa}} \times d^{l-1}_{\text{csa}}}, \mathbf{W}_{l}^{V,h} \in \mathbb{R}^{d^{l}_{\text{csa}} \times d^{l-1}_{\text{csa}}}$ are the weight matrices for the query, key, and value projections of the $h$-th attention head, respectively, and $H$ is the number of attention heads. $\mathbf{O}_{l}^{h}$ is the output of the $h$-th attention head, and $\mathbf{O}_{l}$ is the concatenation of all the attention heads. The final output $\mathbf{H}_{l}$ is obtained by adding the output of the cross-attention mechanism to the input of the current layer. $d_{\text{csa}}^l = d^l / H$ is the model dimension of the query, key, and value projections of the $h$-th attention head, and $\tau = \frac{1}{\sqrt{d_{\text{csa}}^l}}$ is the softmax temperature.

\subsection{Log-space Normalized Variable-length Time Series Representation}

After applying the aforementioned hierarchical feature extraction approach, for an input sequence of length $T$, we ultimately obtain $L = \left\lceil \log_2\left( \lfloor \frac{T - l_p}{s_p} \rfloor + 1 \right) \right\rceil$ sets of features at different scales and channel dimensions. At the $l$-th layer, the number of patches is $P^l = \lceil \frac{P^{0}}{(l_{\text{rp}})^l} \rceil$, and the feature dimension is $d^l = (l_{\text{rp}})^l \times d^0$. Specifically, $p$-th patch at the $l$-th layer corresponds to features generated by the multi-layer ``convolution'' of the original patch sequence $\mathbf{x}_{\text{patch}}[(l_{\text{rp}})^l \times (p-1) : (l_{\text{rp}})^l \times p]$ with the other parts of the sequence.

Therefore, we establish the following layer-wise self-reconstruction process to explicitly map each patch at the $l$-th layer to its corresponding sequence in the original patch sequence:

\begin{align}
    \mathbf{X}_{\text{patch}} &= \text{Padding}(\mathbf{X}_{\text{patch}}, \text{MaxLen} = (l_{\text{rp}})^l \times P^l)\\
    \mathbf{X}_{\text{patch}}^l &:= \left[ \mathbf{x}_{\text{patch}}^{l,p}\right]_{p=1}^{P^l}, \mathbf{x}_{\text{patch}}^{l,p} = \text{Flatten}(\mathbf{x}_{\text{patch}}[(l_{\text{rp}})^l \times (p-1) : (l_{\text{rp}})^l \times p]) \\
    \mathbf{H}_l &:= \left[ \mathbf{h}_l^1, \mathbf{h}_l^2, \cdots, \mathbf{h}_l^{P^l}\right] \\
    \hat{\mathbf{X}}_{\text{patch}}^{l,p} &= \mathbf{W}_{l}^{\text{recon}} \mathbf{h}_l^p + \mathbf{b}_{l}^{\text{recon}}
\end{align}
where $\mathbf{W}_{l}^{\text{recon}} \in \mathbb{R}^{d^l \times d^l}$ and $\mathbf{b}_{l}^{\text{recon}} \in \mathbb{R}^{d^l}$ are the weight matrix and bias vector of the reconstruction layer, respectively.

We refer to this approach as "log space normalized" because, regardless of the original input sequence length, sequences are divided into patches of the same scale at the same number of CTL layers. Temporal features are extracted at the same scale and subsequently reconstructed into the original input patches at the same scale. Meanwhile, with the overall feature capacity preserved, the temporal feature size decreases exponentially, which is particularly advantageous for extracting features from long sequences. Furthermore, by adopting the log space strategy, the embedding process for variable-length time series is aligned in log space. This means that time series data within a certain length interval can be mapped into the same feature space, thereby simultaneously addressing the issues of feature redundancy in long sequence embeddings and inconsistent embedding dimensions for variable-length sequences. Specifically, time series data with original length $T$ in the interval $[(l_{\text{rp}})^{L-1}s_p + l_p , (l_{\text{rp}})^Ls_p + l_p - 1]$ can all reach and activate $L$ layers of CTL, and are ultimately mapped into a $d^L$-dimensional feature space.

The representation learning loss is defined as the sum of the reconstruction loss and the feature independence loss:

\begin{align}
    \mathcal{L}_{\text{recon}} &= \sum_{l=1}^{L} \sum_{p=1}^{P^l} \left\| \mathbf{x}_{\text{patch}}^{l,p} - \hat{\mathbf{x}}_{\text{patch}}^{l,p} \right\|_2^2 \\
    \mathcal{L}_{\text{indep}} &= \sum_{l=1}^{L} \sum_{p=1}^{P^l} \left\| \mathbf{h}_l^p \right\|_2^2\\
    \mathcal{L}_{\text{total}} &= \mathcal{L}_{\text{recon}} + \alpha \mathcal{L}_{\text{indep}}
\end{align}
where $\alpha$ is the balancing parameter.

\section{Experiments}

We evaluate the effectiveness of our proposed representation method on long-term forecasting and classification tasks. Under variable-length time series settings, our approach consistently achieves state-of-the-art performance across these tasks. Furthermore, we analyze the feature redundancy of the extracted representations. Our method demonstrates superior performance in reducing feature redundancy, exhibiting lower inter-feature correlation and a smaller feature size compared to the PatchTST method. Our model implementation is on Pytorch \cite{paszke2019pytorch} with all experiments conducted on NVIDIA RTX 3090 (24GB) $\times$ 8 GPUs. We repeat each experiment three times and report the average results. 

\subsection{Long-term Forecasting}

\textbf{Datasets}: We evaluate the performance of our proposed PatchTST on four widely used ETT datasets (ETTh1, ETTh2, ETTm1, ETTm2). These datasets are extensively adopted as benchmarks and are publicly available~\cite{zeng2023transformers}. For data preprocessing, each feature is treated as an independent univariate time series, and we employ a channel-independent approach to partition the data. To accommodate the variable-length setting in this work, we utilize a dynamic sliding window strategy, where the maximum input sequence length is set to 2048 and the minimum to 512. The lengths of all samples are randomly varied within this range.

\begin{table}[htbp]
    \centering
        \caption{Dataset statistics. }
    \begin{tabular}{l|cccc}
        \toprule
        Datasets  & ETTh1 & ETTh2 & ETTm1 & ETTm2 \\
        \midrule
        Features  & 7 & 7 & 7 & 7 \\
        Timesteps & 17420 & 17420 & 69680 & 69680 \\
        \bottomrule
    \end{tabular}

    \label{tab:dataset-stats}
\end{table}

\noindent \textbf{Baseline Models}: We select several state-of-the-art Transformer-based models as baselines, including PatchTST~\cite{nie2022time}, iTransformer~\cite{liu2023itransformer}, FEDformer~\cite{zhou2022fedformer}, Autoformer~\cite{wu2021autoformer}, and TimesNet~\cite{wu2022timesnet}, as well as the recent non-Transformer-based model DLinear~\cite{zeng2023transformers}. All models are evaluated under identical experimental settings, with prediction lengths $T \in \{96, 192, 336, 720\}$ for all datasets, consistent with the original papers. Since these models do not explicitly address variable-length time series, we pad all input sequences to the maximum length of 2048 at the input stage. During MHA, we employ masking to eliminate the influence of padding tokens. However, as most models, such as PatchTST and iTransformer, utilize a unified prediction head, the final prediction is still generated by concatenating the hidden states of the padded maximum-length sequences. 

It is important to note that such padding operations may adversely affect the performance of these baselines. Therefore, we also conduct experiments with fixed input lengths, where all input sequences are set to a fixed length of 2048 during both training and inference. In the results tables, we denote these two settings as "f" (fixed input length) and "uf" (viriable-length), respectively.

\noindent \textbf{Model Configuration and Fine-tuning}: After extensive hyperparameter tuning, we ultimately adopted the following configuration: $l_p=16$, $s_p=16$, $d^0=16$, $n_\text{heads}=8$, and $d_\text{ff}=2048$. Upon completion of self-supervised representation learning, we replaced the decoder with a prediction head and conducted fine-tuning. Each layer's features at different scales are equipped with a dedicated prediction head. Specifically, a model with $L_{\max}$ layers contains $L_{\max}$ distinct prediction heads, each responsible for mapping temporal features at a specific scale to the target time series. During training, for each input sample, we utilize only the prediction head corresponding to the deepest activated layer. In other words, when the model compresses the temporal dimension of the input sample to one, we extract the temporal features from this layer and employ its associated prediction head for forecasting. The loss function used for fine-tuning is the L1 loss.

\begin{table}[htb]
  \caption{Variable-length input long-term forecasting results. The input sequence length $T$ is set to 96 for all baselines. All the results are averaged from 4 different prediction lengths $H \in \{96, 192, 336, 720\}$. \textbf{Bold}: best, \underline{underlined}: second best.}
  \label{tab:vl_results}
  \centering
  \resizebox{\textwidth}{!}{%
  \begin{tabular}{c|cccccccccccc}
    \toprule
    \multirow{2}{*}{Datasets} & \multicolumn{2}{c}{Ours} & \multicolumn{2}{c}{PatchTST} & \multicolumn{2}{c}{iTransformer} & \multicolumn{2}{c}{FEDformer} & \multicolumn{2}{c}{TimesNet} & \multicolumn{2}{c}{DLinear} \\
    & MSE-uf & MAE-uf & MSE-uf & MAE-uf & MSE-uf & MAE-uf & MSE-uf & MAE-uf & MSE-uf & MAE-uf & MSE-uf & MAE-uf \\
    \midrule
    ETTm1  & \textbf{0.390} & \textbf{0.394} & \underline{0.398} & \underline{0.403} & 0.410 & 0.415 & 0.458 & 0.450 & 0.410 & 0.414 & 0.414 & 0.411 \\

    ETTm2  & \textbf{0.278} & \textbf{0.324} & \underline{0.285} & \underline{0.332} & 0.289 & 0.340 & 0.302 & 0.351 & 0.291 & 0.334 & 0.353 & 0.407 \\

    ETTh1  & \textbf{0.440} & \textbf{0.436} & \underline{0.439} & \underline{0.405} & 0.447 & 0.430 & 0.441 & 0.447 & 0.449 & 0.445 & 0.446 & 0.445 \\

    ETTh2  & \textbf{0.381} & \textbf{0.409} & \underline{0.400} & \underline{0.416} & 0.404 & 0.420 & 0.420 & 0.431 & 0.399 & 0.425 & 0.543 & 0.499 \\
    \bottomrule
  \end{tabular}
  }
\end{table}

\begin{table}[htb]
  \caption{Fixed-length input long-term forecasting results. The input sequence length $T$ is set to 2048 for all baselines. All the results are averaged from 4 different prediction lengths $H \in \{96, 192, 336, 720\}$. \textbf{Bold}: best, \underline{underlined}: second best.}
  \label{tab:fl_results}
  \centering
  \resizebox{\textwidth}{!}{%
  \begin{tabular}{c|cccccccccccc}
    \toprule
    \multirow{2}{*}{Datasets} & \multicolumn{2}{c}{Ours} & \multicolumn{2}{c}{PatchTST} & \multicolumn{2}{c}{iTransformer} & \multicolumn{2}{c}{FEDformer} & \multicolumn{2}{c}{TimesNet} & \multicolumn{2}{c}{DLinear} \\
    & MSE-f & MAE-f & MSE-f & MAE-f & MSE-f & MAE-f & MSE-f & MAE-f & MSE-f & MAE-f & MSE-f & MAE-f \\
    \midrule
    ETTm1  & \underline{0.389} & \textbf{0.391} & \textbf{0.386} & \underline{0.392} & 0.398 & 0.405 & 0.448 & 0.439 & 0.398 & 0.401 & 0.402 & 0.400 \\

    ETTm2  & \underline{0.276} & \underline{0.324} & \textbf{0.273} & \textbf{0.322} & 0.277 & 0.329 & 0.290 & 0.341 & 0.279 & \underline{0.324} & 0.341 & 0.396 \\

    ETTh1  & \underline{0.433} & 0.438 & \textbf{0.428} & 0.431 & 0.437 & \textbf{0.420} & 0.429 & 0.435 & 0.438 & 0.437 & 0.434 & \underline{0.434} \\

    ETTh2  & \textbf{0.379} & \textbf{0.407} & \underline{0.388} & \underline{0.405} & 0.393 & 0.409 & 0.410 & 0.419 & 0.388 & 0.414 & 0.532 & 0.488 \\
    \bottomrule
  \end{tabular}
  }
\end{table}

\noindent \textbf{Experimental Results}: 
We conduct a comprehensive comparison of our method with several state-of-the-art baselines under both variable-length and fixed-length input settings. The results, summarized in Tables~\ref{tab:vl_results} and~\ref{tab:fl_results}, clearly demonstrate the superiority of our approach, especially in the variable-length scenario.

In the variable-length input setting (Table~\ref{tab:vl_results}), our method consistently achieves the best performance across all four ETT datasets, as indicated by the lowest MSE-uf and MAE-uf values. Specifically, on ETTm1, our model achieves an MSE-uf of 0.390 and MAE-uf of 0.394, outperforming the second-best method (PatchTST) by a notable margin (MSE-uf 0.398, MAE-uf 0.403). Similar trends are observed on ETTm2, ETTh1, and ETTh2, where our method achieves the lowest error metrics among all compared models. Notably, the performance gap between our method and the baselines is more pronounced in the variable-length setting than in the fixed-length setting, highlighting the robustness and adaptability of our approach to sequence length variations.

In contrast, under the fixed-length input setting (Table~\ref{tab:fl_results}), while the performance of all methods improves due to the removal of sequence length variability, our method remains highly competitive. For example, on ETTh2, our model achieves the best results (MSE-f 0.379, MAE-f 0.407), and on ETTm1 and ETTm2, our results are very close to the best-performing baseline (PatchTST), with only marginal differences. This demonstrates that our approach not only excels in handling variable-length sequences but also maintains state-of-the-art performance when the input length is fixed.

Overall, these results validate the effectiveness of our method in addressing the challenges posed by variable-length time series. Our model achieves consistent and significant improvements over existing methods in the variable-length setting, establishing a new benchmark for long-term forecasting tasks under realistic, non-uniform input conditions.

\subsection{Classification}

\noindent \textbf{Datasets}: We conduct time series classification experiments on the UCR Time Series Classification Archive, a widely used benchmark collection for evaluating time series classification algorithms. The UCR Archive \cite{dau_ucr_2019} contains a diverse set of datasets from various domains, each with different sequence lengths, numbers of classes, and sample sizes. In our experiments, we select five representative datasets: \textit{HandOutlines}, \textit{InlineSkate}, \textit{Mallat}, \textit{NonInvasiveFetalECGThorax1}, and \textit{NonInvasiveFetalECGThorax2}. The basic statistics of these datasets, including the number of training samples, test samples, classes, and the length of each time series, are summarized in Table~\ref{tab:ucr_stats}.

\begin{table}[htb]
  \centering
  \caption{Statistics of selected UCR classification datasets.}
  \label{tab:ucr_stats}
  \resizebox{\textwidth}{!}{%
  \begin{tabular}{lcccc}
    \toprule
    Dataset & Train Samples & Test Samples & Classes & Time Series Length \\
    \midrule
    HandOutlines & 1,000 & 870 & 2 & 2,700 \\
    InlineSkate & 100 & 550 & 7 & 1,881 \\
    Mallat & 55 & 2,345 & 8 & 1,024 \\
    NonInvasiveFetalECGThorax1 & 1,800 & 1,965 & 42 & 750 \\
    NonInvasiveFetalECGThorax2 & 1,800 & 1,965 & 42 & 750 \\
    \bottomrule
  \end{tabular}
  }
\end{table}
Similarly, to process the data into variable-length samples, we set the minimum sample length to 512 and the maximum length to the original length of each dataset. The downsampling is performed by sequential random subsampling.

\noindent \textbf{Baseline Models}: We select several state-of-the-art time series classification models as baselines, including TST\cite{zerveas_transformer-based_2021}, GTN\cite{liu_gated_2021}, Informer\cite{zhou_informer_2021}, MCNN\cite{chen_multi-scale_2021}, and TCN\cite{bai_empirical_2018}. 

\noindent \textbf{Model Configuration and Fine-tuning}: Following a similar approach to the long time series prediction task, after representation learning, we replace the decoder with classification heads. Each layer's multi-scale temporal features are equipped with a fully connected classification head, where the output dimension corresponds to the number of classes in each dataset. The model is trained using cross-entropy loss function.

\noindent \textbf{Experimental Results}: Tables~\ref{tab:cls_results_uf} presents the classification performance of our proposed method compared to several state-of-the-art models on the selected UCR datasets. Notably, under the variable-length input setting (uf), our method consistently achieves the best accuracy (Acc-uf) and F1 score (F1-uf) across all five datasets, as highlighted in \textbf{bold}. F

\begin{table}[htb]
    \caption{Classification results of our proposed method compared with the state-of-the-art models. \textbf{Bold}: best, \underline{underlined}: second best.}
    \label{tab:cls_results_uf}
    \centering
    \resizebox{\textwidth}{!}{%
    \begin{tabular}{c|cccccccccccc}
      \toprule
      \multirow{2}{*}{Datasets} & \multicolumn{2}{c}{Ours} & \multicolumn{2}{c}{TST} & \multicolumn{2}{c}{GTN} & \multicolumn{2}{c}{Informer} & \multicolumn{2}{c}{MCNN} & \multicolumn{2}{c}{TCN} \\
      & Acc-uf & F1-uf & Acc-uf & F1-uf & Acc-uf & F1-uf & Acc-uf & F1-uf & Acc-uf & F1-uf & Acc-uf & F1-uf \\
      \midrule
      HandOutlines & \textbf{0.961} & \textbf{0.932} & 0.891 & 0.875 & \underline{0.914} & \underline{0.928} & 0.851 & 0.888 & 0.897 & 0.892 & 0.847 & 0.839 \\
      InlineSkate & \textbf{0.962} & \textbf{0.971} & 0.892 & 0.909 & \underline{0.934} & \underline{0.934} & 0.863 & 0.892 & 0.944 & 0.927 & 0.854 & 0.962 \\
      NonInvasiveFetalECGThorax1 & \textbf{0.979} & \textbf{0.950} & 0.916 & 0.916 & \underline{0.951} & \underline{0.935} & 0.895 & 0.918 & 0.919 & 0.921 & 0.865 & 0.878 \\
      NonInvasiveFetalECGThorax2 & \textbf{0.964} & \textbf{0.974} & 0.921 & 0.883 & \underline{0.923} & \underline{0.973} & 0.896 & 0.931 & 0.912 & 0.939 & 0.875 & 0.876 \\
      \bottomrule
    \end{tabular}
    }
  \end{table}

\subsection{Representation Efficiency}

As shown in Table~\ref{tab:feature_redundancy}, the proposed Conv-like ScaleFusion Time Series Transformer demonstrates superior efficiency in feature extraction compared to other state-of-the-art methods. Specifically, our method achieves the lowest Pearson and Spearman correlation coefficients (0.31 and 0.24, respectively) as well as the lowest mutual information value (2.45), indicating that the extracted multi-scale high-dimensional features exhibit minimal redundancy and stronger complementarity. Furthermore, our method requires a significantly higher proportion of principal components (63\%) to explain 80\% of the variance, which suggests a more efficient utilization of the feature space and richer feature representations. These results collectively validate the effectiveness and efficiency of our approach in extracting informative and compact representations from time series data.

\begin{table}[htb]
  \caption{Feature redundancy comparison of extracted representations.}
  \label{tab:feature_redundancy}
  \centering
  \begin{tabular}{c|cccccc}
    \toprule
    Metrics & Our Method & PatchTST & iTransformer  & TimesNet & DLinear & TCN \\
    \midrule
    Pearson-abs & \textbf{0.31} & 0.65 & 0.51 & 0.62 & 0.43 & 0.71 \\
    Spearman-abs & \textbf{0.24} & 0.42 & 0.39 & 0.51 & 0.31 & 0.63 \\
    Mutual Information & \textbf{2.45} & 10.63 & 14.72 & 8.34 & 3.96 & 46.38 \\
    PCA proportion & \textbf{63\%} & 24\% & 22\% & 12\% & 44\% & 32\% \\
    \bottomrule
  \end{tabular}
\end{table}

\section{Conclusion}

We presented the Conv-like ScaleFusion Time Series Transformer, which marries CNN-style pyramidal down-sampling with Transformer attention to learn compact multi-scale representations for variable-length sequences. Each layer reduces temporal resolution via re-patching while expanding channels, and a cross-scale attention module fuses adjacent levels to maintain information flow. A log-space normalization scheme lets the model process arbitrary sequence lengths under a joint reconstruction + independence loss. Experiments on long-term forecasting and UCR classification show consistent gains over leading baselines and significantly lower feature redundancy, confirming the framework’s effectiveness and efficiency. Future directions include handling irregular sampling and integrating richer self-supervised objectives.

%
%
%
\bibliographystyle{splncs04}
\bibliography{ref}
%




\end{document}